\documentclass[runningheads]{llncs}
\usepackage[T1]{fontenc}
\usepackage{amsmath,amssymb,amsfonts}
\usepackage{booktabs}
\usepackage{graphicx}
\usepackage{url}
\usepackage{multirow}
\usepackage{color}

\begin{document}
\title{Adaptive Conformal Prediction for Reliable and Explainable Medical Image Classification}
\titlerunning{Adaptive Conformal Prediction for Medical Image Classification}
%

\author{One Octadion\orcidID{0009-0008-6823-2816} \and 
Novanto Yudistira\orcidID{0000-0001-5330-5930} \and 
Lailil Muflikhah\orcidID{0000-0001-7903-0576}}

\institute{Faculty of Computer Science, Universitas Brawijaya, Malang, Indonesia\\
\email{octadion@student.ub.ac.id, \{yudistira,lailil\}@ub.ac.id}}

\maketitle
\begin{abstract}
Deep learning models for medical imaging often exhibit overconfidence, creating safety risks in ambiguous diagnostic scenarios. While Conformal Prediction (CP) provides distribution-free statistical guarantees, standard methods like Regularized Adaptive Prediction Sets (RAPS) optimize for average efficiency, masking severe failures on difficult inputs. We propose an Adaptive Lambda Criterion for RAPS that minimizes the worst-case coverage violation across prediction set size strata. On OrganAMNIST (58,850 abdominal CT images, 11~classes), standard size-optimized RAPS converges to near-deterministic behavior with stratified undercoverage on uncertain samples, while our method achieves 95.72\% global coverage with average set size 1.09 and $\ge$90\% coverage across all strata. Cross-domain validation on PathMNIST (107,180 pathology images, 9~classes) confirms generalizability. Quantitative Grad-CAM analysis ($\rho=-0.30$, $p<10^{-22}$) confirms that multi-label predictions correspond to focused attention on anatomically ambiguous regions.

\keywords{Medical Image Classification \and Conformal Prediction \and Uncertainty Quantification \and Stratified Coverage \and Explainable AI}
\end{abstract}
\section{Introduction}

The integration of Deep Learning (DL) into radiological workflows promises to revolutionize diagnostic efficiency, particularly in automated analysis of abdominal Computed Tomography (CT) scans~\cite{yang2023medmnist}. However, translation to clinical practice is stalled by a lack of reliable uncertainty quantification. Standard classifiers are frequently overconfident, assigning high probabilities even to ambiguous or misclassified inputs~\cite{guo2017calibration}. In high-stakes medical settings, a model that is ``mostly correct'' but confidently wrong on difficult cases violates the core tenet of safety.

Conformal Prediction (CP) has emerged as a rigorous framework to address this, transforming model outputs into prediction sets guaranteed to contain the true label with high probability~\cite{angelopoulos2021gentle}. Yet, applying standard CP methods to medical imaging reveals a dangerous limitation. Our analysis on OrganAMNIST shows that while Regularized Adaptive Prediction Sets (RAPS)~\cite{angelopoulos2020uncertainty} meet global coverage targets, they suffer from stratified undercoverage: standard RAPS produces prediction sets of size $\ge 2$ for only 10 out of 17,778 test samples, with coverage on these uncertain samples dropping to 60\%. This ``safety gap'' emerges exactly when diagnostic uncertainty is highest.

\begin{figure}[t]
\centering
\includegraphics[width=\textwidth]{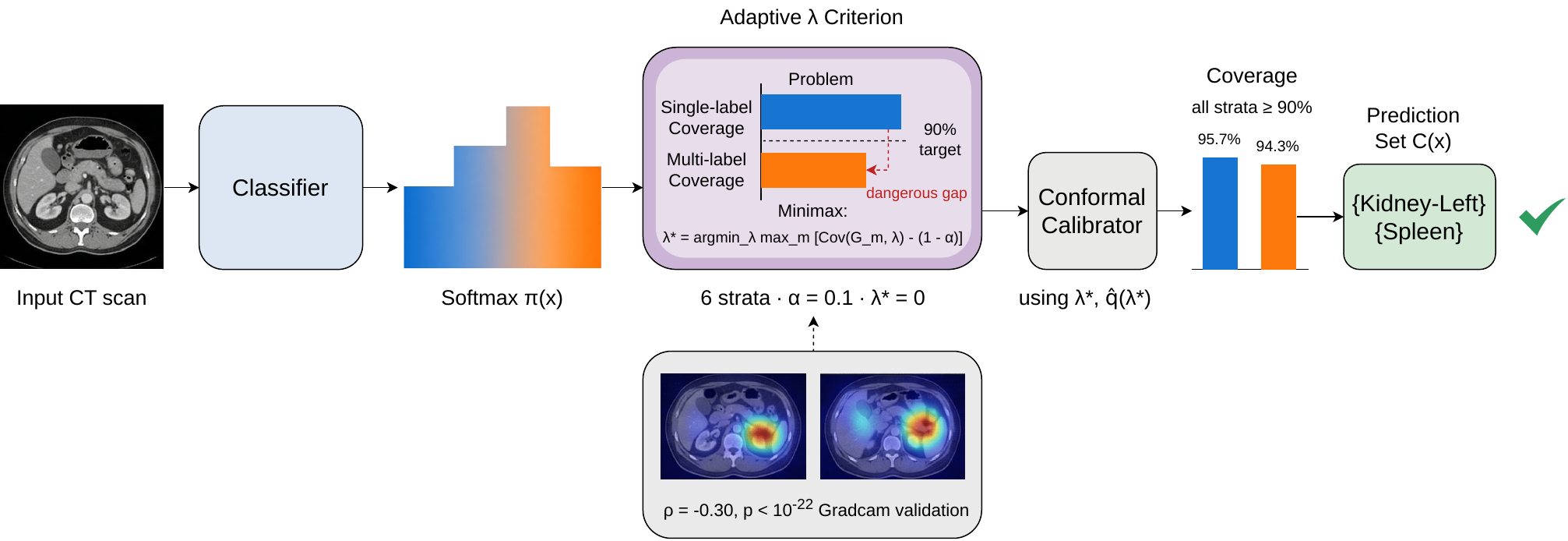}
\caption{System overview. Our framework wraps a ResNet-18 backbone with an Adaptive Conformal Prediction layer. By optimizing $\lambda$ via stratified minimax on a calibration set, the system generates dynamic prediction sets (e.g., \{Kidney-Left, Spleen\}) guaranteed to contain the true label with 90\% probability.}
\label{fig:teaser}
\end{figure}

We propose a novel Adaptive Lambda Criterion for RAPS (Fig.~\ref{fig:teaser}) that minimizes the worst-case violation across uncertainty strata. Our method restores coverage to $\ge$90\% for difficult cases while maintaining high efficiency (average set size 1.09), validated on both OrganAMNIST and PathMNIST. We further integrate Grad-CAM~\cite{selvaraju2017grad} to provide qualitative and quantitative validation that expanded prediction sets are semantically grounded.

The main contributions are:
\begin{itemize}
    \item We expose a critical failure mode of size-optimized RAPS on ambiguous medical images, where stratified coverage collapses despite satisfying marginal guarantees.
    \item We introduce a stratified minimax objective (\textit{Adaptive Lambda Criterion}) ensuring uniform coverage across all uncertainty levels.
    \item We validate on two imaging modalities---OrganAMNIST (CT) and PathMNIST (pathology)---demonstrating cross-domain generalizability.
    \item We provide quantitative Grad-CAM analysis confirming that prediction set expansion reflects genuine diagnostic ambiguity.
\end{itemize}

The remainder of this paper is organized as follows. Section~\ref{sec:related} reviews related work. Section~\ref{sec:methods} details our methodology. Section~\ref{sec:experiments} presents experimental results on OrganAMNIST and PathMNIST. Section~\ref{sec:discussion} discusses clinical implications, theoretical considerations, and limitations.

\section{Related Work}\label{sec:related}

\subsection{Uncertainty Quantification in Medical AI}
Predictive uncertainty estimation is a fundamental challenge across machine learning~\cite{tyralis2024review}, with particular urgency in medical applications where prediction errors carry life-threatening consequences. Deep learning has been widely applied to medical prediction tasks spanning radiology, genomics~\cite{muflikhah2024snp}, and disease prognosis~\cite{sreenivasan2025conformal}, yet standard models trained with cross-entropy loss tend to force single predictions even in ambiguous cases, leading to overconfidence~\cite{guo2017calibration}.

For medical image classification specifically, CNNs like ResNet~\cite{he2016deep} serve as backbones but struggle with epistemic ambiguity in low-resolution benchmarks~\cite{yang2023medmnist}, particularly bilateral organ symmetry. Bayesian approaches including MC~Dropout~\cite{gal2016dropout} and Deep Ensembles~\cite{lakshminarayanan2017simple} improve calibration but incur high computational costs or require multiple models. Post-hoc methods like Temperature Scaling~\cite{guo2017calibration} improve probability reliability but lack individual safety guarantees~\cite{angelopoulos2021gentle}.

\subsection{Conformal Prediction Methodologies}
Conformal Prediction (CP) provides distribution-free statistical guarantees on prediction sets~\cite{vovk2005algorithmic}, making it a principled framework for safety-critical applications. For a comprehensive survey, see Zhou et al.~\cite{zhou2025conformal}. Early classification approaches like LAC~\cite{saddler1999comparison} used fixed thresholds but failed to adapt to data heteroscedasticity. Romano et al.~\cite{romano2020classification} introduced Adaptive Prediction Sets (APS), accumulating probability mass until reaching target confidence. To address the ``long tail'' issue, Angelopoulos et al.~\cite{angelopoulos2020uncertainty} proposed RAPS, adding a penalty $\lambda$ for larger sets---currently the state-of-the-art for multi-class tasks. Seedat et al.~\cite{seedat2023sscp} further improved adaptive CP using self-supervised learning.

Recent applications demonstrate CP's versatility across domains: genomic medicine~\cite{lu2022reliable}, graph neural networks~\cite{zargarbashi2023conformal}, semantic segmentation~\cite{bojer2025conformal}, and conformal triage for clinical deployment~\cite{fisch2022conformal}. However, most works apply standard CP directly without addressing stratified failures on difficult subgroups.

\subsection{The Gap: Stratified Coverage}
Standard CP coverage guarantees are \emph{marginal} (averaged over populations), allowing global validity while systematically failing on difficult subgroups. Recent works highlight the need for size-stratified~\cite{angelopoulos2021gentle,ding2024rank} and class-conditional coverage~\cite{ding2024rank}. Gauthier et al.~\cite{gauthier2025adaptive} formalize adaptive coverage policies, and Babbar et al.~\cite{babbar2022utility} demonstrate the utility of well-calibrated prediction sets in human--AI collaboration. Our work addresses this gap with an Adaptive Lambda criterion ensuring uniform coverage across uncertainty levels, preventing safety compromises on difficult diagnostic cases.

\section{Methodology}\label{sec:methods}

\subsection{Problem Formulation}
We consider a multi-class classification task with input space $\mathcal{X} \subseteq \mathbb{R}^{H \times W \times C}$ and label space $\mathcal{Y} = \{0, \dots, K{-}1\}$. Given a dataset $\mathcal{D} = \{(X_i, Y_i)\}_{i=1}^n$ drawn i.i.d.\ from an unknown joint distribution $P_{XY}$, we construct a set-valued predictor $\mathcal{C}: \mathcal{X} \rightarrow 2^\mathcal{Y}$ satisfying the marginal coverage guarantee:
\begin{equation}
    \mathbb{P}(Y \in \mathcal{C}(X)) \ge 1 - \alpha
\end{equation}
where $1{-}\alpha$ is the target coverage level (e.g., $0.9$ for $\alpha{=}0.1$).

\subsection{Base Classifier Architecture}
Our base model $f_\theta$ is a ResNet-18~\cite{he2016deep} pretrained on ImageNet-1K. The final fully connected layer is replaced with a Dropout layer ($p{=}0.3$) followed by a linear projection to $\mathbb{R}^{K}$ (where $K{=}11$ for OrganAMNIST, $K{=}9$ for PathMNIST). The model is trained to minimize Cross-Entropy loss with label smoothing ($\epsilon{=}0.1$) to prevent overfitting and encourage cluster tightness. Optimization uses the Adam optimizer with a learning rate of $0.001$ and cosine annealing schedule. We employ early stopping with patience of 10~epochs based on validation accuracy.

\subsection{Conformal Prediction Framework}
We adopt Split Conformal Prediction (SCP). The validation data is partitioned into a tuning set $\mathcal{D}_{\text{tune}}$ (for hyperparameter selection) and a calibration set $\mathcal{D}_{\text{cal}}$ (for quantile computation).

\subsubsection{Regularized Adaptive Prediction Sets (RAPS).}
We utilize RAPS~\cite{angelopoulos2020uncertainty}, which modifies the conformal score by penalizing large prediction sets. Let $\pi(x)$ be the softmax output. We sort probabilities such that $\pi_{(1)}(x) \ge \pi_{(2)}(x) \ge \dots \ge \pi_{(K)}(x)$. The cumulative mass for the true label $y$ at rank $k$ is:
\begin{equation}
    S_{\text{base}}(x,y) = \sum_{j=1}^{k} \pi_{(j)}(x) \quad \text{where } \pi_{(k)}(x) = \pi_y(x)
\end{equation}
RAPS augments this with a regularization penalty:
\begin{equation}
    S_{\text{RAPS}}(x,y) = S_{\text{base}}(x,y) + \lambda \max(0, k - k_{\text{reg}})
\end{equation}
where $k_{\text{reg}}$ is a rank threshold (automatically tuned or fixed) and $\lambda \ge 0$ penalizes lower-ranked classes, effectively truncating the ``tail'' of the probability distribution.

\subsubsection{Proposed: Adaptive Lambda Optimization.}
Standard RAPS selects $\lambda$ to minimize average prediction set size (\textbf{Size Criterion}), which often leads to poor coverage on ``hard'' samples requiring larger sets.

We propose the \textbf{Adaptive Lambda Criterion}. We define $M{=}6$ disjoint strata based on prediction set cardinality: $G_1{=}\{0,1\}$ (empty or singleton), $G_2{=}\{2,3\}$ (low ambiguity), $G_3{=}\{4,5,6\}$ (medium ambiguity), $G_4{=}\{7,\dots,10\}$, $G_5{=}\{11,\dots,100\}$, and $G_6{=}\{101,\dots,K'\}$. The upper strata are effectively empty for $K \le 11$ but ensure generalizability to larger label spaces. Ablation studies (Section~\ref{sec:ablation}) confirm robustness to the choice of strata boundaries.

We formulate the selection of $\lambda$ as a minimax optimization problem over a discretized grid $\Lambda_{\text{grid}}$:
\begin{equation}
    \lambda^*_{\text{adaptive}} = \mathop{\arg\min}_{\lambda \in \Lambda_{\text{grid}}} \left( \max_{m=1 \dots M} \left| \text{Cov}(G_m, \lambda) - (1{-}\alpha) \right| \right)
\end{equation}
where $\text{Cov}(G_m, \lambda)$ is the empirical coverage on $\mathcal{D}_{\text{tune}}$ for the subset of images whose predicted set falls into stratum $G_m$. Based on our empirical analysis, we use $\Lambda_{\text{grid}} = \{0, 10^{-5}, 10^{-4}, 8{\times}10^{-4}, 9{\times}10^{-4}, 10^{-3}, 1.5{\times}10^{-3}, 2{\times}10^{-3}\}$. This objective ensures that the model does not sacrifice safety on difficult inputs to achieve efficiency on easy ones.

\subsection{Experimental Setup}

\paragraph{Datasets.}
We evaluate on two MedMNIST benchmarks~\cite{yang2023medmnist}: (1)~\textbf{OrganAMNIST}: 58,850 abdominal CT images across 11 organ classes ($28{\times}28$ grayscale, upsampled to $224{\times}224$), split into 34,581/6,491/17,778 for train/val/test; and (2)~\textbf{PathMNIST}: 107,180 colon pathology images across 9 tissue types ($28{\times}28$ RGB), split into 89,996/10,004/7,180. These represent fundamentally different imaging modalities (radiology vs.\ histopathology) and ambiguity structures.

\paragraph{Baselines.}
We compare: (1)~\textbf{Naive}: softmax thresholding ($\pi_k(x) \ge 1{-}\alpha$); (2)~\textbf{LAC}: label-conditional adaptive CP~\cite{saddler1999comparison}; (3)~\textbf{RAPS~(Size)}: $\lambda$ optimized for minimal average set size, without temperature scaling ($T{=}1$); (4)~\textbf{RAPS~(Temp)}: RAPS~(Size) with temperature scaling ($T$ optimized via SGD on $\mathcal{D}_{\text{cal}}$); and (5)~\textbf{RAPS~(Adaptive)}: our stratified minimax criterion.

\paragraph{Metrics.}
All experiments use $\alpha{=}0.1$ (90\% target coverage). We report: (1)~\textbf{Marginal Coverage}: global validity check; (2)~\textbf{Average Set Size (APSS)}: efficiency metric (lower is better); (3)~\textbf{Singleton Rate}: proportion of single-label outputs; (4)~\textbf{Empty Rate}: proportion of empty prediction sets; and (5)~\textbf{Strat.\ Min.}: the worst-case coverage across all populated prediction set size strata, which is the key safety indicator.

\section{Experimental Results}\label{sec:experiments}

\subsection{Base Classifier Performance}
Before applying conformal prediction, we evaluated the underlying ResNet-18 classifier. On OrganAMNIST, the model achieves 93.20\% accuracy. Table~\ref{tab:class_perf} reveals substantial heterogeneity across organ classes.

\begin{table}[t]
\caption{Per-class performance of the base ResNet-18 classifier on OrganAMNIST. Bold highlights the class with the highest epistemic uncertainty due to bilateral symmetry.}
\label{tab:class_perf}
\centering
\begin{tabular*}{\textwidth}{@{\extracolsep{\fill}}lcccc@{}}
\toprule
Class & Precision & Recall & F1 & Support \\
\midrule
Bladder & 0.934 & 0.878 & 0.906 & 1,036 \\
Femur-Left & 0.971 & 0.903 & 0.936 & 784 \\
Femur-Right & 0.901 & 0.951 & 0.925 & 793 \\
Heart & 1.000 & 0.876 & 0.934 & 785 \\
\textbf{Kidney-Left} & \textbf{0.712} & \textbf{0.936} & \textbf{0.809} & \textbf{2,064} \\
Kidney-Right & 0.957 & 0.784 & 0.862 & 1,965 \\
Liver & 0.997 & 0.980 & 0.988 & 3,285 \\
Lung-Left & 1.000 & 0.996 & 0.998 & 1,747 \\
Lung-Right & 0.978 & 0.991 & 0.984 & 1,813 \\
Pancreas & 0.963 & 0.983 & 0.973 & 1,622 \\
Spleen & 0.957 & 0.895 & 0.925 & 1,884 \\
\bottomrule
\end{tabular*}
\end{table}

Organs with distinct morphology (Lungs, Liver) achieve near-perfect classification (F1$>$0.98). In contrast, ``Kidney-Left'' exhibits the lowest precision (71.2\%) due to bilateral confusion with ``Kidney-Right'' and anatomical proximity to ``Spleen''. This confusion pattern---where the classifier absorbs false positives from two structurally similar classes---is a central motivation for uncertainty quantification that adapts to per-class difficulty. On PathMNIST, the classifier achieves 91.77\% accuracy (macro-F1: 0.896).

\subsection{Comparative Analysis of Conformal Methods}
Table~\ref{tab:uq_metrics} compares all five methods on OrganAMNIST. The critical column is Strat.\ Min., which reveals whether coverage holds uniformly across prediction set size strata.

\begin{table}[t]
\caption{Conformal Prediction methods on OrganAMNIST ($\alpha{=}0.1$). Strat.\ Min.: worst-case coverage across populated strata. $\dagger$~Only one populated stratum.}
\label{tab:uq_metrics}
\centering
\begin{tabular*}{\textwidth}{@{\extracolsep{\fill}}lccccc@{}}
\toprule
Method & Coverage & Avg Size & Singleton & Empty & Strat.\ Min.\ \\
\midrule
Naive$^\dagger$ & 0.9320 & 1.00 & 100.0\% & 0.0\% & 0.932$^\dagger$ \\
LAC$^\dagger$ & 0.8051 & 0.82 & 82.0\% & 18.0\% & 0.805$^\dagger$ \\
RAPS (Size) & 0.9321 & 1.00 & 99.9\% & 0.0\% & 0.600 \\
RAPS (Temp) & 0.9321 & 1.00 & 99.9\% & 0.0\% & 0.400 \\
\textbf{RAPS (Adaptive)} & \textbf{0.9572} & \textbf{1.09} & 93.0\% & 0.0\% & \textbf{0.943} \\
\bottomrule
\end{tabular*}
\end{table}

The Naive method satisfies coverage (93.20\%) solely because top-1 accuracy is high, but cannot expand sets for uncertain inputs. LAC fails entirely (80.51\% coverage, 18.0\% empty sets). Both RAPS~(Size) and RAPS~(Temp) achieve 93.21\% marginal coverage but converge to near-deterministic behavior, producing only 10 multi-label sets out of 17,778. The Strat.\ Min.\ column exposes the safety gap: coverage on these few predictions drops to 60\% and 40\%, respectively. RAPS~(Adaptive) achieves 95.72\% global coverage with average set size~1.09, generating 1,244 multi-label predictions (7\%) while maintaining $\ge$94\% coverage across all populated strata.

\subsection{Stratified Coverage Analysis}
\begin{figure}[t]
\centering
\includegraphics[width=0.9\textwidth]{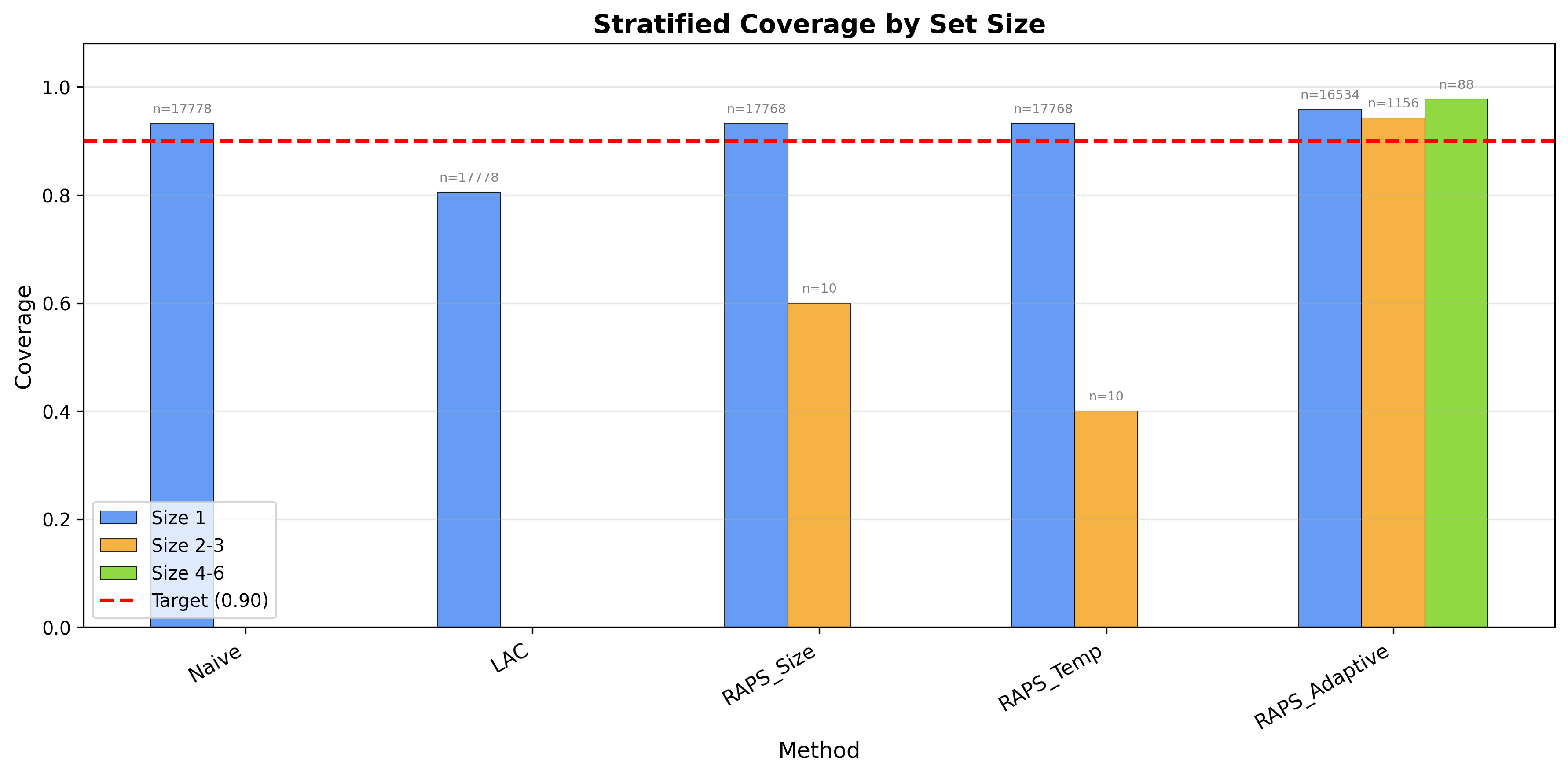}
\caption{Stratified coverage on OrganAMNIST. RAPS~(Size) produces only 10 multi-label sets with 60\% coverage; RAPS~(Adaptive) maintains $\ge$90\% across all populated strata.}
\label{fig:stratified}
\end{figure}

Global metrics hide local failures. Fig.~\ref{fig:stratified} visualizes the core finding.

Standard RAPS~(Size) achieves 93.23\% on singletons but only 60\% on 10 multi-label predictions. RAPS~(Adaptive) produces a meaningful distribution (964 pairs, 192 triples, 88 sets of size $\ge$4) with 95.81\% ($G_1$), 94.29\% ($G_2$; $n{=}1{,}156$), and 97.73\% ($G_3$) coverage.

\subsection{Cross-Domain Validation on PathMNIST}
To assess generalizability, we replicated the full pipeline on PathMNIST (colon pathology, 9~classes, $N_{\text{test}}{=}7{,}180$). Table~\ref{tab:pathmnist} summarizes the results.

\begin{table}[t]
\caption{Conformal Prediction results on PathMNIST ($\alpha{=}0.1$). The same failure pattern and remedy observed on OrganAMNIST generalize to a different imaging modality.}
\label{tab:pathmnist}
\centering
\begin{tabular*}{\textwidth}{@{\extracolsep{\fill}}lccccc@{}}
\toprule
Method & Coverage & Avg Size & Singleton & Empty & Strat.\ Min.\ \\
\midrule
Naive$^\dagger$ & 0.9177 & 1.00 & 100.0\% & 0.0\% & 0.918$^\dagger$ \\
LAC$^\dagger$ & 0.7299 & 0.74 & 74.3\% & 25.7\% & 0.730$^\dagger$ \\
RAPS (Size) & 0.9182 & 1.00 & 99.9\% & 0.0\% & 0.750 \\
RAPS (Temp) & 0.9181 & 1.00 & 99.9\% & 0.0\% & 0.778 \\
\textbf{RAPS (Adaptive)} & \textbf{0.9500} & \textbf{1.12} & 91.2\% & 0.0\% & \textbf{0.881} \\
\bottomrule
\end{tabular*}
\end{table}

The same failure pattern emerges: standard RAPS produces only 8 multi-label predictions with 75\% stratum coverage. RAPS~(Adaptive) generates 634 multi-label sets (8.8\%) with 95.00\% global coverage and $\ge$88\% on all populated strata, confirming cross-domain generalizability.

\subsection{Ablation Studies}\label{sec:ablation}
We assess sensitivity of RAPS~(Adaptive) to two key design choices. Table~\ref{tab:ablation} reports results on OrganAMNIST.

\begin{table}[t]
\caption{Ablation study on OrganAMNIST. The Adaptive Lambda method is robust to both strata granularity and lambda grid resolution. All configurations select $\lambda{=}0$, $k_{\text{reg}}{=}1$.}
\label{tab:ablation}
\centering
\begin{tabular*}{\textwidth}{@{\extracolsep{\fill}}llccc@{}}
\toprule
Experiment & Configuration & Coverage & Avg Size & Strat.\ Min.\ \\
\midrule
\multirow{3}{*}{Strata} & Coarse (3 strata) & 0.9582 & 1.097 & 0.9399 \\
& Default (6 strata) & 0.9592 & 1.103 & 0.9447 \\
& Fine (8 strata) & 0.9582 & 1.101 & 0.9469 \\
\midrule
\multirow{3}{*}{$\Lambda_{\text{grid}}$} & Coarse (5 pts) & 0.9601 & 1.106 & 0.9446 \\
& Default (8 pts) & 0.9586 & 1.100 & 0.9388 \\
& Fine (12 pts) & 0.9592 & 1.104 & 0.9425 \\
\bottomrule
\end{tabular*}
\end{table}

\paragraph{Strata Boundaries.} Three configurations---coarse (3~strata), default (6~strata), and fine (8~strata)---all select the same optimal hyperparameters ($\lambda{=}0$, $k_{\text{reg}}{=}1$). Worst-case stratum coverage varies within a narrow range (93.99\%--94.69\%) and average set sizes remain between 1.097 and 1.103, indicating robustness to strata granularity.

\paragraph{Lambda Grid Resolution.} Grids of 5, 8, and 12~points spanning $[0, 0.01]$ yield consistent results: worst-case stratum coverage ranges from 93.88\% to 94.46\%, with no monotonic improvement from refinement. The default 8-point grid provides sufficient resolution.

\subsection{Explainability with Grad-CAM}\label{sec:gradcam}
To validate that expanded prediction sets are semantically grounded, we integrate Gradient-weighted Class Activation Mapping (Grad-CAM)~\cite{selvaraju2017grad} with both qualitative and quantitative analysis.

\begin{figure}[t]
\centering
\includegraphics[width=\textwidth]{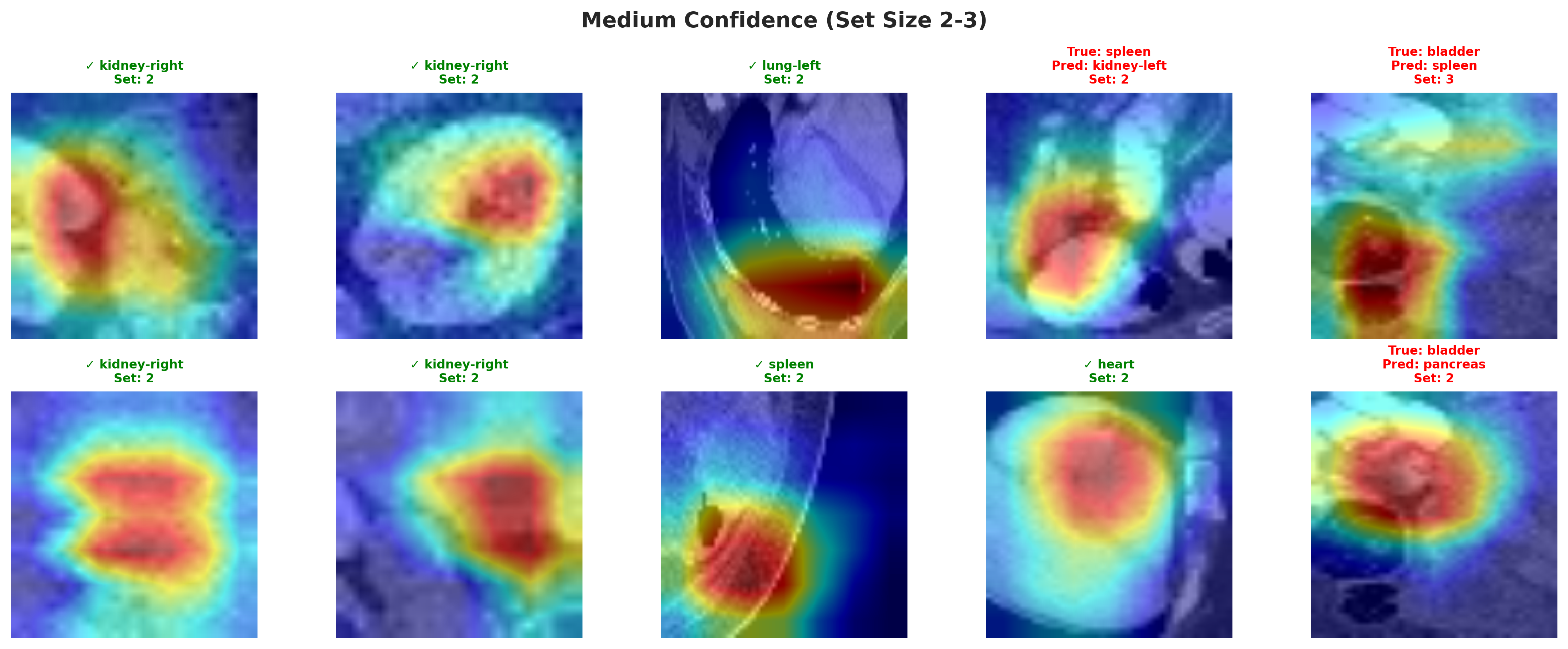}
\caption{Grad-CAM visualizations for multi-label predictions (set size 2--3). Heatmaps reveal attention concentrated on organ boundaries and low-contrast regions, confirming that expanded prediction sets reflect genuine anatomical ambiguity.}
\label{fig:gradcam_medium}
\end{figure}

Fig.~\ref{fig:gradcam_medium} shows representative samples where RAPS~(Adaptive) produces sets of size $\ge$2. The Grad-CAM heatmaps consistently highlight organ boundaries, bilateral structures, and low-contrast regions, confirming that the model expands its prediction set when attending to anatomically ambiguous features.

\paragraph{Quantitative Analysis.} We compute the spatial entropy of each Grad-CAM heatmap as a measure of attention dispersion and correlate it with prediction set size over 1,000 test samples. Table~\ref{tab:gradcam} summarizes the results.

\begin{table}[t]
\caption{Quantitative Grad-CAM analysis on OrganAMNIST ($n{=}1{,}000$). Spatial entropy is significantly lower (more focused attention) for multi-label predictions, confirming semantic grounding.}
\label{tab:gradcam}
\centering
\begin{tabular*}{\textwidth}{@{\extracolsep{\fill}}lcc@{}}
\toprule
Measure & Value & $p$-value \\
\midrule
Spearman $\rho$ (entropy vs.\ set size) & $-0.303$ & $< 10^{-22}$ \\
Point-biserial $r$ (entropy vs.\ singleton) & $-0.438$ & $< 10^{-48}$ \\
\midrule
Mean entropy (singleton predictions) & \multicolumn{2}{c}{15.37} \\
Mean entropy (multi-label predictions) & \multicolumn{2}{c}{15.16} \\
\bottomrule
\end{tabular*}
\end{table}

Spearman rank correlation yields $\rho = -0.303$ ($p < 10^{-22}$): samples with larger prediction sets exhibit more spatially concentrated Grad-CAM activations. Point-biserial correlation ($r = -0.438$, $p < 10^{-48}$) and the entropy gap between singletons (15.37) and multi-label predictions (15.16) confirm that model uncertainty is systematically associated with focused attention on ambiguous regions.

\begin{figure}[t]
\centering
\includegraphics[width=\textwidth]{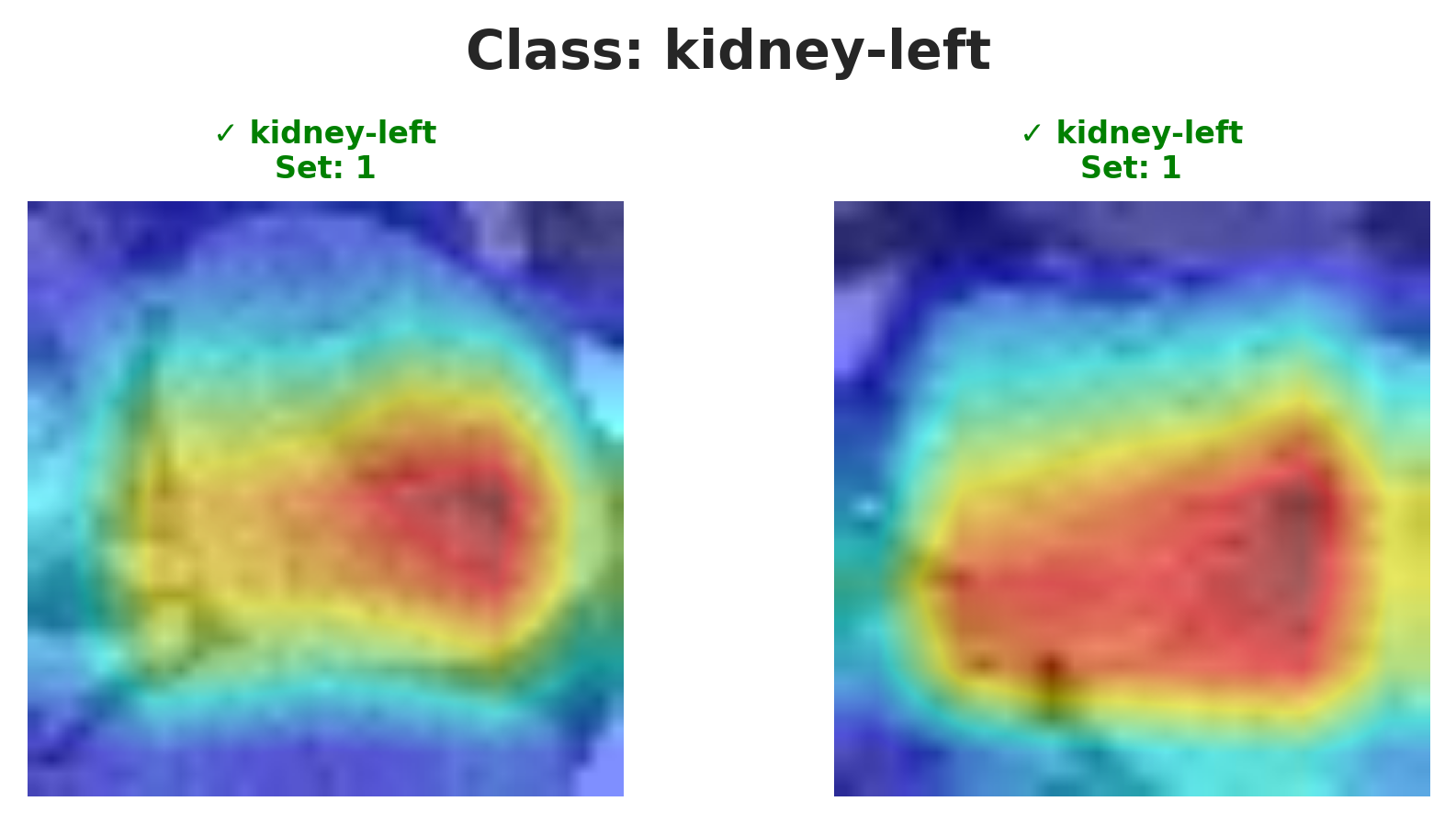}
\caption{Grad-CAM for ``Kidney-Left'', the class with the lowest precision (71.2\%, Table~\ref{tab:class_perf}). Attention frequently bleeds into the spleen and contralateral kidney regions, explaining the bilateral confusion pattern.}
\label{fig:gradcam_kidney}
\end{figure}

Fig.~\ref{fig:gradcam_kidney} provides a per-class view for ``Kidney-Left'', revealing attention extending into the spleen and contralateral kidney---directly explaining the low precision in Table~\ref{tab:class_perf}.

\section{Discussion}\label{sec:discussion}

\subsection{Safety--Efficiency Trade-off and Bilateral Ambiguity}
Our experiments reveal that standard RAPS converges to near-deterministic behavior by selecting an excessively aggressive $\lambda$, dominated by the ``easy'' majority. For ambiguous samples, coverage collapses to 40--60\% (Tables~\ref{tab:uq_metrics}--\ref{tab:pathmnist}). The Adaptive Lambda criterion restores stratum-level coverage to $\ge$90\% at a modest cost of ${\sim}$9\% increase in average set size. The quantitative Grad-CAM analysis (Section~\ref{sec:gradcam}, $\rho{=}{-}0.30$, $p{<}10^{-22}$) confirms that expanded prediction sets reflect focused attention on anatomically ambiguous regions---particularly bilateral organ boundaries (Table~\ref{tab:class_perf})---effectively mimicking a radiologist's ``differential diagnosis'' process.

\subsection{Theoretical Considerations}
Distribution-free stratum-level coverage guarantees are fundamentally impossible without additional assumptions on the data-generating process~\cite{lei2018distribution}; standard CP only provides marginal guarantees over the entire test distribution~\cite{gauthier2025adaptive}. Our Adaptive Lambda criterion is therefore a practical safety heuristic that empirically minimizes worst-case stratified violations. The consistent improvements across both datasets suggest the minimax strategy is an effective remedy for safety-critical deployments.

\subsection{Clinical Implications and Limitations}
From a regulatory perspective regarding Software as a Medical Device (SaMD)~\cite{fda2021samd}, uniformity of performance is paramount. Well-calibrated prediction sets can meaningfully improve human--AI decision-making~\cite{babbar2022utility}. The Adaptive Lambda framework aligns with these requirements by ensuring the ``90\% probability of inclusion'' applies to every patient.

Both benchmarks use low-resolution images ($28{\times}28$); some observed uncertainty may be artifactual. Validation on full-resolution clinical data remains important future work. Extending the minimax criterion to class-conditional and intersectional fairness is another promising direction.

\begin{credits}
\subsubsection{\discintname}
The authors have no competing interests to declare that are relevant to the content of this article.
\end{credits}

\end{document}